\documentclass{article}

\usepackage[preprint]{neurips_2022}

\usepackage[utf8]{inputenc} 
\usepackage[T1]{fontenc}    
\usepackage{hyperref}       
\usepackage{url}            
\usepackage{booktabs}       
\usepackage{amsfonts}       
\usepackage{nicefrac}       
\usepackage{microtype}      
\usepackage{xcolor}         

\usepackage{amsmath,amssymb} 
\usepackage{graphicx}
\usepackage{color}
\usepackage{multirow}
\usepackage{float}

\setcitestyle{numbers,open={[},close={]}} 

\title{Permutation-Invariant Relational Network for Multi-person 3D Pose Estimation}

\author{%
   Nicolas Ugrinovic \\
    IRI\\
   CSIC-UPC \\
   08028, Barcelona, Spain \\
   \texttt{first.last@upc.edu} \\
   \And
   Adria Ruiz \\
   Seedtag\\
   08014, Barcelona, Spain\ \\
   \texttt{adriaruiz@seedtag.com} \\
   \And
   Antonio Agudo \\
       IRI\\
   CSIC-UPC \\
   08028, Barcelona, Spain \\
   \texttt{first.last@upc.edu} \\
   \And
   Alberto Sanfeliu \\
    IRI\\
   CSIC-UPC \\
   08028, Barcelona, Spain \\
   \texttt{first.last@upc.edu} \\
   \And
   Francesc Moreno-Noguer \\
    IRI\\
   CSIC-UPC \\
   08028, Barcelona, Spain \\
   \texttt{first.last@upc.edu} \\
}

\begin{document}

\newcommand{\GA}[1]{{\color{violet}#1}}
\newcommand{\GPM}[1]{{\color{blue} GPM:#1}} 

\newcommand{\mA}{\mathcal{A}}
\newcommand{\mF}{\mathcal{F}}
\newcommand{\mI}{\mathcal{I}}
\newcommand{\mK}{\mathcal{K}}
\newcommand{\mP}{\mathcal{P}}
\newcommand{\mR}{\mathcal{R}}
\newcommand{\mcU}{\mathcal{U}}
\newcommand{\mZ}{\mathcal{Z}}

\renewcommand{\vec}[1]{\boldsymbol{#1}}
\newcommand{\mat}[1]{\mathbf{#1}}
\newcommand{\set}[1]{\mathcal{#1}}

\newcommand{\real}[0]{\mathbb{R}}
\newcommand{\tb}[0]{\textbf}
\newcommand{\ti}[0]{\textit}
\newcommand{\et}[0]{\ti{et al.}}

\newcommand{\imageset}[0]{\set{I}}
\newcommand{\image}[0]{\mat{I}}

\newcommand{\poseset}[0]{\set{P}}
\newcommand{\transset}[0]{\set{T}}
\newcommand{\jointset}[0]{\set{J}}
\newcommand{\garmparamset}[0]{\set{G}}

\newcommand{\template}[0]{\mat{T}}
\newcommand{\garment}[0]{\mat{G}}
\newcommand{\blendweight}[0]{w}
\newcommand{\blendweights}[0]{\mat{W}}

\newcommand{\normal}[0]{{\mathbf{n}}}

\newcommand{\depth}[0]{\widehat{\vec{d}}}
\newcommand{\ankl}[0]{\mathbf{x}_{l}}
\newcommand{\ankr}[0]{\mathbf{x}_{r}}
\newcommand{\lplane}[0]{L_{p}}
\newcommand{\lamdp}[0]{\lambda_{p}}

\newcommand{\pose}[0]{\vec{\theta}}
\newcommand{\shape}[0]{\vec{\beta}}

\newcommand{\joints}[0]{\mat{J}}
\newcommand{\jointsTwoD}[0]{\tilde{\mat{J}}}

\newcommand{\rot}[0]{\vec{R}}

\newcommand{\scale}[0]{\mat{s}}
\newcommand{\trans}[0]{\vec{t}}

\newcommand{\nplane}[0]{\bf{\widehat{{n}}} }

\newcommand{\garmparam}[0]{\vec{z}}
\newcommand{\offsets}[0]{\mathbf{D}}

\newcommand{\cut}{\mat{z}_\mathrm{cut}}
\newcommand{\style}{\mat{z}_\mathrm{style}}
\newcommand{\posenc}{\mat{P}_{\shape}}
\newcommand{\zpose}{\mat{z}_{\pose}}

\newcommand{\smpl}[0]{M}
\newcommand{\posefun}[0]{T}
\newcommand{\blendfun}[0]{W}
\newcommand{\offsetfun}[0]{B}
\newcommand{\offsetsfun}[0]{D}
\newcommand{\jointfun}[0]{J}
\newcommand{\garmfun}[0]{G}

\newcommand{\metricdist}{pairwise normalized distances between persons}

\newcommand{\numberOfMetrics}{three} 

\newcommand{\Modelname}{Keep Your Feet on the Ground} 

\newcommand{\lossRep}[0]{L}
\newcommand{\multilossRep}[0]{\hat{L}}
\newcommand{\lambdaRep}[0]{\lambda_{2D}}

\newcommand{\heightDistMetric}[0]{h_{err}}

\newcommand{\posevect}[0]{\vec{p}}
\newcommand{\posevectRef}[0]{\vec{q}}
\newcommand{\jointsDims}[0]{\mathbb{R}^{J\times3}}
\newcommand{\diffVect}[0]{\vec{\delta}}

\newcommand{\featureVect}[0]{\vec{f}}
\newcommand{\featureNet}[0]{\mathcal{G}}
\newcommand{\ourNet}[0]{\Phi}
\newcommand{\rfineNet}[0]{\mathcal{R}}

\newcommand{\reals}[0]{\mathbb{R}}

\maketitle

\begin{abstract}

The recovery of multi-person 3D poses from a single RGB image is a severely ill-conditioned problem due to the inherent 2D-3D depth ambiguity, inter-person occlusions, and body truncations. To tackle these issues, recent works have shown promising results by simultaneously reasoning for different people. However, in most cases this is done by only considering pairwise person interactions,  hindering thus a holistic scene representation able to capture long range interactions. This is addressed by approaches that jointly process all people in the scene, although they require defining one of the individuals as a reference and a pre-defined person ordering, being sensitive to this choice. In this paper, we overcome both these limitations, and we propose an approach for multi-person 3D pose estimation that captures long range interactions independently of the input order. For this purpose we build a residual-like permutation-invariant network that successfully refines potentially corrupted initial 3D poses estimated by an off-the-shelf detector. The residual function is learned via {\em Set Transformer}~\cite{set_transformer_lee2019} blocks, that model the interactions among all initial poses, no matter their ordering nor number. 
A thorough evaluation demonstrates that our approach is able to boost the performance of the initially estimated 3D poses by large margins, achieving state-of-the-art results on standardized benchmarks. Additionally, the proposed module works in a computationally efficient manner and can be potentially used as a drop-in complement for any 3D pose detector in multi-people scenes. 

\end{abstract}

\section{Introduction}


Estimating 3D human pose from RGB images is a long-standing problem in computer vision, with broad applications in, e.g., action recognition, 3D content production, the game industry and  human-robot interaction. While recent approaches have shown impressive results, most of them focus on a single person~\cite{li20143d,mehta2017vnect,Zhou_2017_ICCV,pavlakos2017coarse,pavlakos_learning_2018,Sun_2018_ECCV,Zhao_2019_CVPR,Martinez_2017_ICCV,zhou2016sparseness}. The problem of multi-person 3D pose estimation introduces additional challenges to the single person setup, mostly due to inter-person occlusions. In order to tackle it, sequential~\cite{zanfir_cvpr_2018_multiple,TDBU_cheng_2021_CVPR} and multi-camera systems~\cite{Dong2019,Lin2021,Tu2020,Wu2021} have been exploited. In this paper, however, we aim to address the most constricting version of the problem by considering multi-people 3D pose estimation from one single view. 



Only a reduced number of works have proposed  strategies to better exploit multi-person relations. For instance,~\cite{HMOR2020} and ~\cite{jiang2020_multiperson} use depth-aware ordinal losses to better estimate the pose of each person in relation to others. Nevertheless, these methods still reason about multiple people in a local neighborhood and in a strictly pairwise manner. They do not consider the fact that human interactions can occur at large ranges and among several people in the scene. For instance, in sports, a team formation is correlated as a whole and not at local regions of the pitch. Leveraging on a self-attention layer, Pi-Net~\cite{guo2021pi} introduced a mechanism to simultaneously reason for all people in the image. However, this approach is sensitive to permutations in the input order as it also uses an RNN which are known to have this short-comming~\cite{vinyals2015order}. Fieraru~\et~\cite{Fieraru_2021_NEURIPS} capture people interactions with specialized losses for a variable number of people. However, they use a parametric human body model and rely on direct supervision for modeling closed contact human interactions. 
Moreover, at the time of this writing, there does not exist any publicly available dataset with human-human interactions annotations; these being difficult to produce. 

\begin{figure*}[!t]
  \vspace{-0.45cm}
  \centering
  \includegraphics[width=0.9\linewidth, trim={0.1cm 0.1cm 0cm 0}, clip = true]{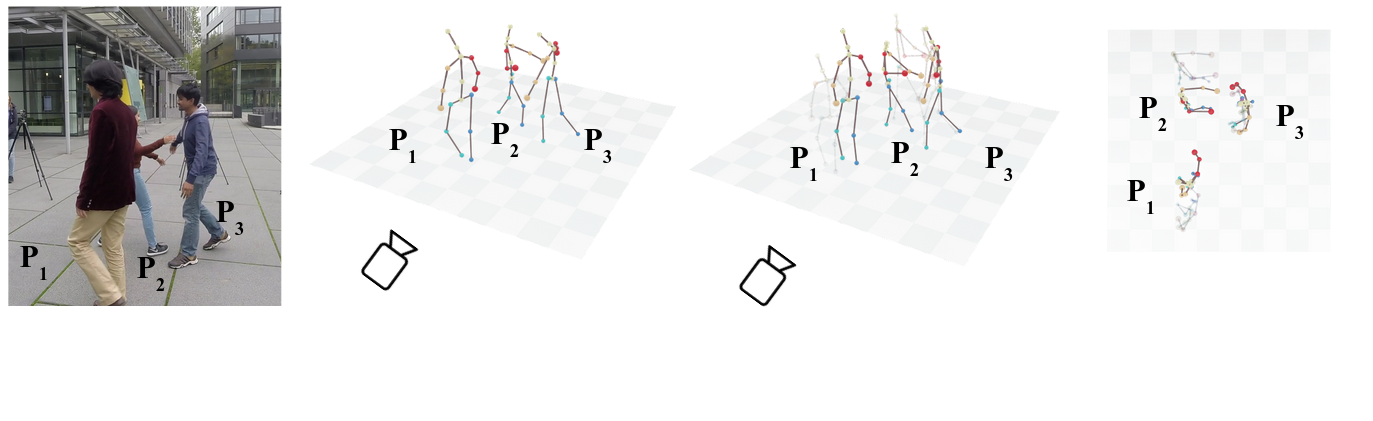}
          \put(-340,20){{{Input image}}}
        \put(-255,20){{{Ground truth}}}
        \put(-170,20){{{Refined (and Initial)}}}
        \put(-58,25){{{Bird-view}}}
        \put(-80,15){{{(Refined and Initial)}}}
  \vspace{-0.6cm}
  \caption{\small{Given a set of potentially noisy  input 3D poses, we leverage on the Set Transformer architecture~\cite{set_transformer_lee2019} to compute a holistic encoding of all poses. This encoding which can take an arbitrarily large number of poses in any order, helps to predict a residual for each pose and refine the initial estimates. The approach is robust to large errors on the initial poses. Note how our refinement corrects the scale and translation of person P\textsubscript{2}. Our model also shows improvements in the root-relative pose (see main text).}}
  \label{fig:teaser}
  \vspace{-0.4cm}
\end{figure*}
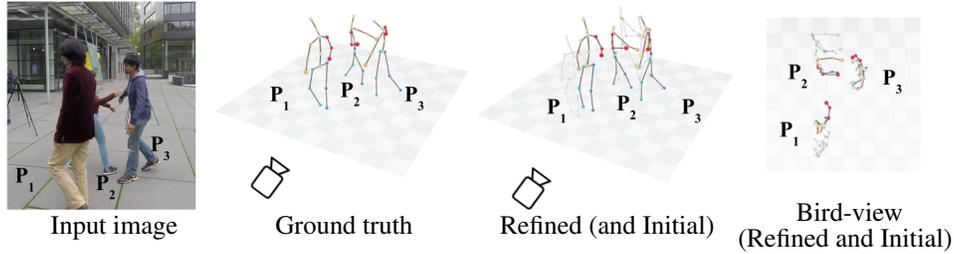

In order to overcome the limitations of previous approaches, we propose a novel scheme to model people interactions in a holistic and permutation-invariant fashion. In our approach, we assume access to an initial set of potentially noisy 3D poses estimated by an off-the-shelf algorithm~\cite{Moon_2019_ICCV_3DMPPE}. These poses are treated as elements of a set and fed into a Set Transformer-based architecture~\cite{set_transformer_lee2019}. This allows us to simultaneously process all of input poses, and compute a global encoding not sensitive to the input order. This encoding contains interaction information and is then used to estimate a correction residual for the initial 3D poses. Note, that we do not train with direct interaction supervision. Instead, our model learns interactions between person's poses implicitly.
Figure~\ref{fig:teaser} shows how our model refines the initial pose estimations and yields to more correct ones, even under strong occlusions. This way, capturing the interactions between input poses, the model can correct the relative pose, translation, and scale of the people in the scene.

A thorough evaluation on MuPoTS-3D~\cite{singleshot}, Panoptic~\cite{panoptic} and NBA2K~\cite{zhu_2020_eccv_nba} multi-people datasets shows that the proposed approach consistently improves over Pi-Net~\cite{guo2021pi} and other state-of-the-art (SOTA) methods~\cite{HMOR2020,smap_eccv2020}. This happens even when initial 3D poses estimated by the off-the-shelf~\cite{Moon_2019_ICCV_3DMPPE} network are noisy, corrupted by occlusions or contain truncated body joints. 


Interestingly, the proposed approach runs efficiently and can be potentially used as a post-processing module on top of any pose estimation method with negligible computation overhead. In summary, our key contributions are the following:
(1) We introduce a novel approach to capture the relationship among the 3D poses of multiple people.
(2) Our model does not depend on the input order (i.e., it is permutation-invariant) and can handle an arbitrarily large number of people. 
(3) The proposed module is computationally efficient and could be used along with any 3D pose detector.

\section{Related work}\label{sec:sota}


In this paper, we focus on \textbf{single-view multi-person 3D human pose estimation}. In recent years there has been a growing attention on this topic. Typically, there exist two types of approaches: top-down~\cite{lcr,lcr++,Moon_2019_ICCV_3DMPPE,hdnet_2020,darval_multi_2019,zanfir_cvpr_2018_multiple,jiang2020_multiperson} and bottom-up~\cite{mehta2017vnect,singleshot,mehta2017monocular,zanfir_multipeople_nips18,xnect_mehta_2020,smap_eccv2020,ROMP,BMP_2021_CVPR}. Although, very recently, a new trend to integrate both approaches has emerged~\cite{HMOR2020,TDBU_cheng_2021_CVPR,khirodkar2022occluded}. In this work, we use
a top-down approach~\cite{Moon_2019_ICCV_3DMPPE} exploiting contextual information, specifically, human-human interactions.


The idea of using \textbf{human-human interaction} information to improve 3D human pose estimation of multiple people was first proposed by~\cite{andriluka2012human_context}. However, it was not until recently that the community shifted its attention to exploit this information. Recently, Jiang~\et~\cite{jiang2020_multiperson} and~\cite{HMOR2020}  exploit depth-order relationships between people. \cite{TDBU_cheng_2021_CVPR} propose a pose discriminator to capture two-person natural interactions. \cite{Fieraru_2020_CVPR} proposed to exploit close contact information between two persons. Finally, closest to our approach, Guo~\et~\cite{guo2021pi} and Fieraru~\et~\cite{Fieraru_2021_NEURIPS} propose pose interacting networks to exploit human-to-human interaction information. 
In this paper, we aim to capture the interaction of people present in a scene in a  \textbf{permutation-invariant} manner, meaning that the order we input each person to our model should not matter. Ideally, we should be able to treat all the poses as elements of a set. 
\cite{santoro2017simple} proposes a relational network that sum-pools all pairwise interactions of elements in a given set. However, we want to model higher-order interactions as the pose of a person may affect (or be affected by) not one but multiple other persons directly or indirectly. At the same time, a group of people could affect a person's pose. \cite{ma2018attend} uses a Transformer~\cite{attention_is_all_2017} to model high-order interactions between objects. However, they use mean-pooling to obtain aggregated features where interaction information may be loss. More suited to our problem, we choose~\cite{set_transformer_lee2019} as our base architecture.

\begin{figure*}[t!]
  \vspace{-1.0cm}
  \centering
  \includegraphics[width=\linewidth, scale=0.7,trim={0.1cm 0.1cm 0cm 0}, clip = true]{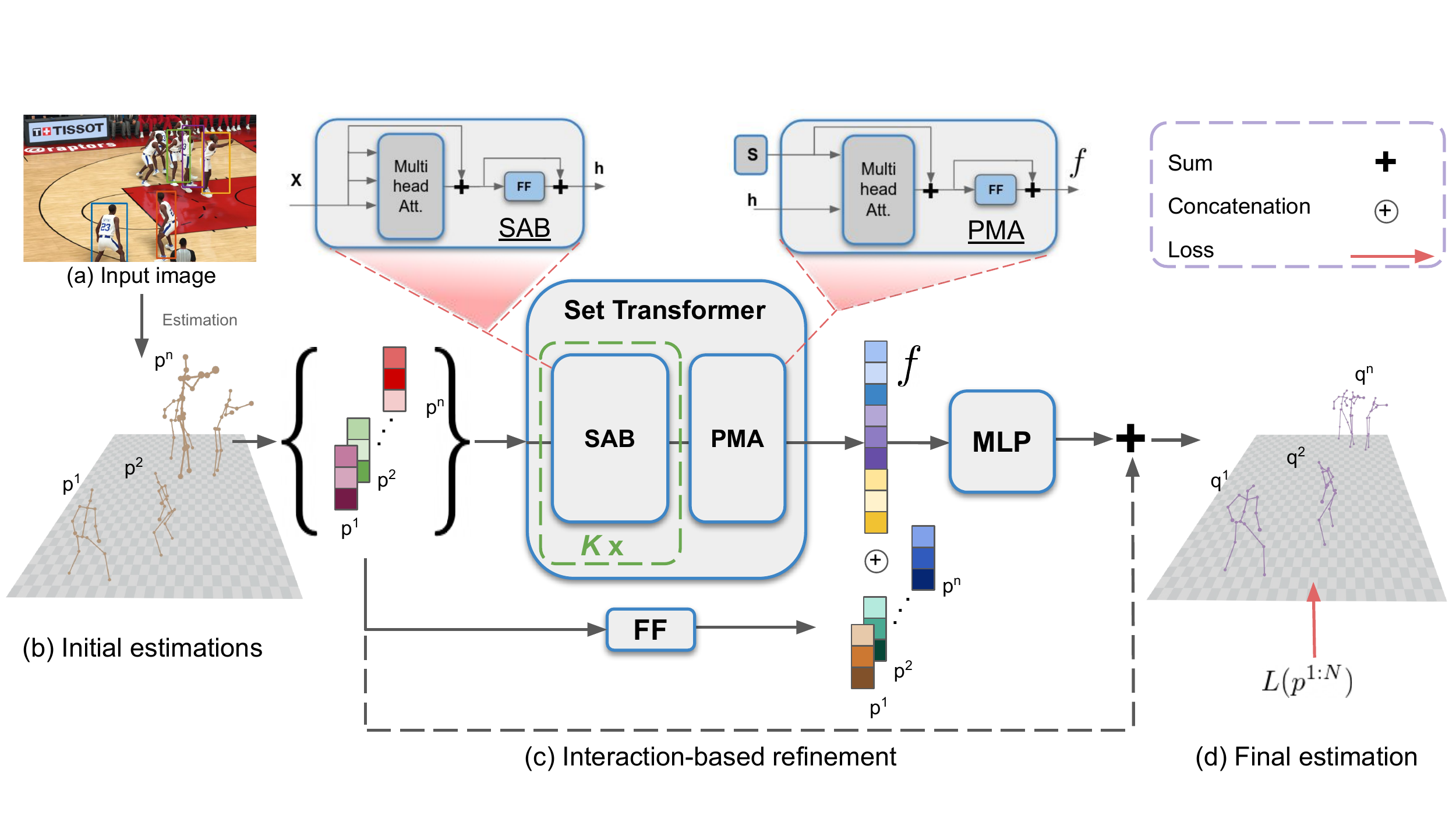}
  \vspace{-0.5cm}
  \caption{\small{\textbf{Overview of our approach.} Given an input image (a), we first estimate the 3D keypoints as the initialization (b). Then, we input these initial estimations in the form of a set (hence the keys) to our interaction-based permutation-invariant model. We obtain the \ti{interaction-based embedding} ($f$) and concatenate it with another embedding for each person. This \ti{person} embedding is calculated as a projection from the input space to the same dimension as $f$ via a feed-forward layer, denoted FF. We use an MLP network to get the corrections of the initial estimations (b) and compute the final estimations (d) by adding these corrections to the initial poses. We show the poses of the people with bounding boxes in the image just for clarity. Our model inputs all the poses in the scene. 
  }}
  \label{fig:model}
  \vspace{0.1cm}
\end{figure*}

\section{Method}\label{sec:method}

The key idea of our approach is to implicitly capture the interaction information between all human body poses in a scene and use it to refine an initial set of 3D pose estimations. We represent this information with an \textit{interaction embedding}. People in a scene are usually involved in a specific interaction and this constrains the range of possible poses. Thus, we argue that learning pose correlations can improve initial noisy estimations. Inspired by the effectiveness of Transformers~\cite{attention_is_all_2017} to capture correlations, we use the Set Transformer~\cite{set_transformer_lee2019}, a model unaffected by the order of its inputs. In short, we obtain an information-rich and permutation-invariant interaction embedding that captures human-to-human interactions and use it to refine each person's pose.

\subsection{Relational Network for Multi-person 3D Pose Estimation} 
\label{subsec:formulation}

Let ${\bf I} \in \mathbb{R}^{H \times W \times 3}$ be an input RGB image with $N$ people interacting in the same scene and ${\posevect}^{1:N}$ to be the set of 3D joints corresponding to each person where $\posevect^{n} \in \jointsDims$ with $J$ number of estimated joints and $n$ a number in the range $\{1,...,N\}$. These joints are obtained from an initial estimation using an off-the-shelf 3D pose estimator, such as~\cite{Moon_2019_ICCV_3DMPPE}. All $N$ poses are assumed to be represented in absolute camera coordinates.

Given the previous definitions, we aim to improve each initial pose estimation~$\posevect^{n}$ taking into account the pose of all the people present in the scene. These initial estimations could be inaccurate as they are sensitive to inter-person occlusions, self-occlusions, scale/depth ambiguity, and truncation. Although the latter does not originate from a lack of interaction information, we shall see that our model also deals with these cases. This is because, aside from modeling interaction information, our model also captures the error distribution of the initial estimation method. 

The mapping between these initial estimations and the refined poses takes the form $\posevectRef^{1:N}=\ourNet(\posevect^{1:N})$,
where $\posevectRef^{1:N}$ refers to the set of poses $\{\posevectRef^{1},...,\posevectRef^{n}\}$ refined by exploiting the interactions, and $\posevectRef^n \in \jointsDims$. The function $\ourNet$ materializes as a neural-network capable of extracting interaction information between the input poses~$\posevect^{1:N}$ and refining them. 


For this purpose, we split the problem into two parts: First, we obtain an embedding (\textit{interaction embedding}) able to capture the interactions of the scene; second, we use this to refine the initial poses. The \textit{interaction embedding}~$\featureVect$ is a $d$-dimensional vector where $d$ is a hyperparameter of our model. The embedding~$\featureVect$, obtained via  Set Transformer blocks, aims to globally capture interactions between people from the initial estimations~$\posevect^{1:N}$. Specifically, 
\begin{equation} 
    \featureVect =\featureNet(\posevect^{1:N}),
    \label{eq:feature}
\end{equation}
where $\featureNet$ is a neural network composed of the Set Transformer elements described in the next section. Ideally, we are looking for a function $\featureNet$ capable of capturing the embedding regardless the input order and the number of people in the scene. If we explicitly take into account the \textit{interaction embedding}, we can express the relation between the initial and refined estimations as:
\begin{equation} 
    \posevectRef^{1:N}=\ourNet(\featureVect,\posevect^{1:N}),
    \label{eq:goal}
\end{equation}
where~$\ourNet$ is our full model described in more detail in Sec.~\ref{subsec:model} and depicted in Fig.~\ref{fig:model}.

\subsection{Computing interaction embeddings with Set Transformers}
\label{subsec:set_transformer}

As stated before, our \textit{interaction embedding}~$\featureVect$ should comply with two key requirements to model the interaction information: (1) being independent of the order of the input  person's body joints and (2) being able to process input scenes containing any number of people. Requirement (1) comes from the fact that we do not want the input order of our model to affect the pose refinement. Only the information regarding interaction between people's body poses should affect the refinement. Both requirements are not easily satisfied by classical feed-forward neural networks, and recursive neural networks (RNNs) are sensitive to the input order~\cite{vinyals2015order}.
Thus, to get the desired interaction embedding, we base our model in an attention-based permutation-invariant neural network. For this purpose, we take components from the Set Transformer~\cite{set_transformer_lee2019}. 
Choosing an attention-based architecture for our model and working with sets as inputs allows us to: compute a variable number of input poses, disregard the input order, and naturally encode interaction between these elements. In this manner, we are able to attend to any person's initial pose and obtain a rich permutation-invariant embedding that captures the interactions in the scene. We then use this feature to guide the pose refinement process.  


In our context, we treat the initial pose estimations~$\posevect^{1:N}$ as a set ((b) in Fig.~\ref{fig:model}). Thus, our model attends to each person's joints and first generates an embedding for each person using a Set Attention Block (SAB). Later, these individual embeddings are aggregated in a learned fashion using a Pooling by Multihead Attention (PMA) operation, providing us the interaction embedding~$\featureVect$. For a formal definition of the SAB and PMA modules, we refer the reader to~\cite{set_transformer_lee2019}.




The SAB module used here emerges as an adaptation of the encoder block of the Transformer~\cite{attention_is_all_2017}. To build the SAB, dropout and the positional encoding are discarded. This module uses self-attention to concurrently encode the input set. This allows to capture pairwise and higher-order relationships among instances during the encoding process. The output of the SAB contains information about pairwise interactions among the elements of the input set $\vec{X}$ and can be stacked $K$ times by more of the same modules to capture higher than pairwise interactions. In our context, $\vec{X}$ is composed by the set of initial pose estimations~$\posevect^{1:N}$, thus, $\vec{X} = \{\posevect^{1},...,\posevect^{n}\}$. 

To obtain a permutation-invariant feature, we use the Pooling by Multihead Attention (PMA) operation which aggregates the features obtained by the SAB. This constitutes a key step to make~$\featureVect$ permutation-invariant. The PMA operation aggregates the features by applying multi-head attention on a learnable set of $k$ seed vectors $\vec{S} \in \reals^{k \times d}$. In our case, $k=1$, as we only have one embedding to represent the whole scene. 
At the output of the \textrm{PMA} block, we find the desired \textit{interaction embedding}. Under the definitions provided before, we define our embedding in the following manner:
  \begin{equation} 
    \featureVect = \textrm{PMA}_1(\textrm{SAB}(\vec{X}))\;.
    \label{eq:embedding_pma}
\end{equation}



\subsection{Pose refinement via interaction information}
\label{subsec:model}
Once we have the feature $\featureVect$, we use them to refine the initial 3D human pose estimations. For this, we employ an MLP that takes as input  $\featureVect$ concatenated with a projection of each person's joints into a $d$-dimensional vector. This projection is done via a feed-forward layer, denoted FF. Finally, the MLP outputs a vector ($\diffVect$) containing all the correction values needed to improve each of the initial joints locations in 3D space. 
We define correction vector as:
\begin{equation} 
    \diffVect^{1:N}=\ourNet(\featureVect,\posevect^{1:N}),
    \label{eq:delta}
\end{equation}
where our network $\ourNet$ is based on the SAB/PMA modules and the MLP in charge of decoding the interaction embedding. The set-processing modules (SAB and PMA) generate the interaction embedding, as considered in Eq.~(\eqref{eq:feature}). Adding this correction vector to our initial estimations, we can now compute the refined joints by the following relation:
\begin{equation} 
    \posevectRef^{1:N}=\posevect^{1:N} + \diffVect^{1:N}\;.
    \label{eq:refine_net}
\end{equation}
To guide the learning process, we optimize the whole network parameters
by minimizing an $L_2$ loss over the final refined 3D joints and the ground truth joints:
\begin{equation} 
    L(\posevect^{1:N})=\frac{1}{N}\sum_{i=1}^{N}||\posevectRef^i - \posevect_{GT}^i||^2,
    \label{eq:refine_net2}
\end{equation}
where $\posevect_{GT}$ denotes the 3D human pose ground truth.
For more details regarding our final architecture, please refer to the supplementary material and the code of this paper.

\section{Experiments}\label{sec:experiments}

In this section, we present implementation details, the evaluation of our approach in comparison with other relevant SOTA, and an ablation study focusing on various types of interactions. Finally, we present an analysis of the refinement process and the computational complexity of our model. Our model has the advantage of being highly computationally efficient, lightweight and fast to train (see Sec~\ref{sec:complexity}). 
We experiment on three datasets: MuPoTS-3D~\cite{singleshot}, Panoptic~\cite{panoptic} and NBA2K~\cite{zhu_2020_eccv_nba}. Additionally, we include qualitative results on COCO~\cite{coco}. We also use standard metrics for evaluation such as \textbf{MPJPE}~\cite{ionescu_human3.6m:_2014} --which measures the accuracy of the 3D root-relative pose-- and \textbf{3DPCK}~\cite{mehta_monocular_2016} with a threshold of 15cm, as it is standard in the literature~\cite{guo2021pi,smap_eccv2020,hdnet_2020}. Complementary to 3DPCK (from now on PCK), we use \textbf{AUC} (area under the curve) as a more complete metric. Additionally, \textbf{PCK\textsubscript{abs}} is used to evaluate absolute camera-centered 3D human poses.


\subsection{Implementation details}
\label{sec:implementation}
We optimize our model parameters using ADAM~\cite{adam} with learning rate of 0.0001 in a single GTX 1080 Ti. To estimate the initial 3D human poses we use~\cite{Moon_2019_ICCV_3DMPPE}. Although, any other initialization approach could be used. Regarding training data, although generally used for this task, we discard the use of MuCo-3DHP. Given its synthetic nature it does not contain real interactions between people. Instead, we use \textbf{MuPoTS-3D} as both train and test set which does contain real interactions. For fair comparison, we resource to k-fold cross-validation dividing the dataset into 10 folds. For evaluating on the \textbf{Panoptic} studio~\cite{panoptic} dataset, we follow the evaluation protocol presented in~\cite{zanfir_cvpr_2018_multiple,zanfir_multipeople_nips18}. Finally, please notice that even though the \textbf{NBA2K dataset} is synthetic, it captures plausible interaction between players as opposed to MuCo-3DHP. For more details regarding data and implementation, please refer to the supplementary material.

\subsection{Comparison with state-of-the-art methods}

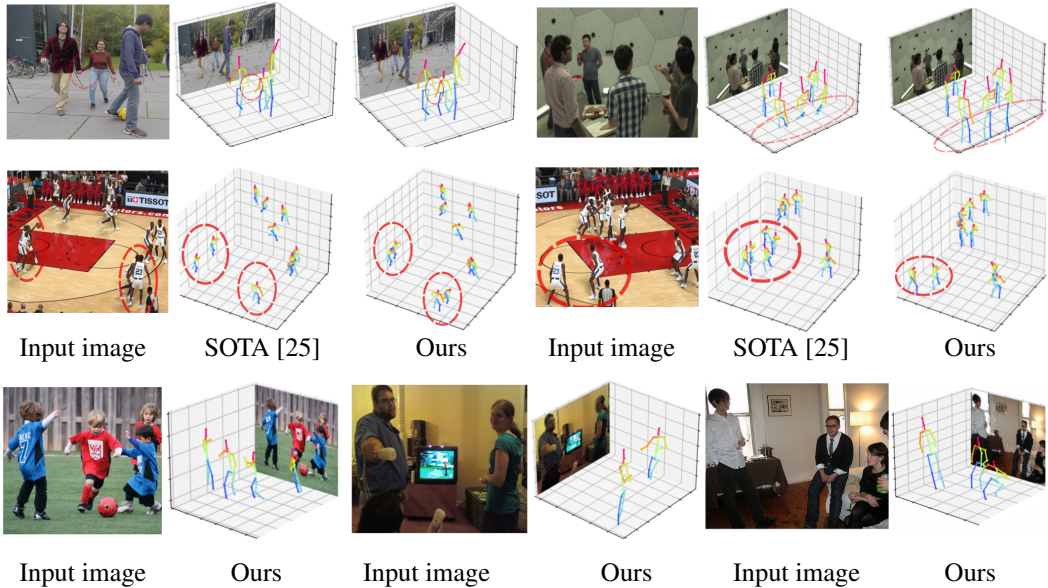
\begin{figure*}[t!]
  \vspace{-0.1cm}
    \centering
  \includegraphics[width=\linewidth, trim={0.1cm 0.1cm 0cm 0}, clip = true]{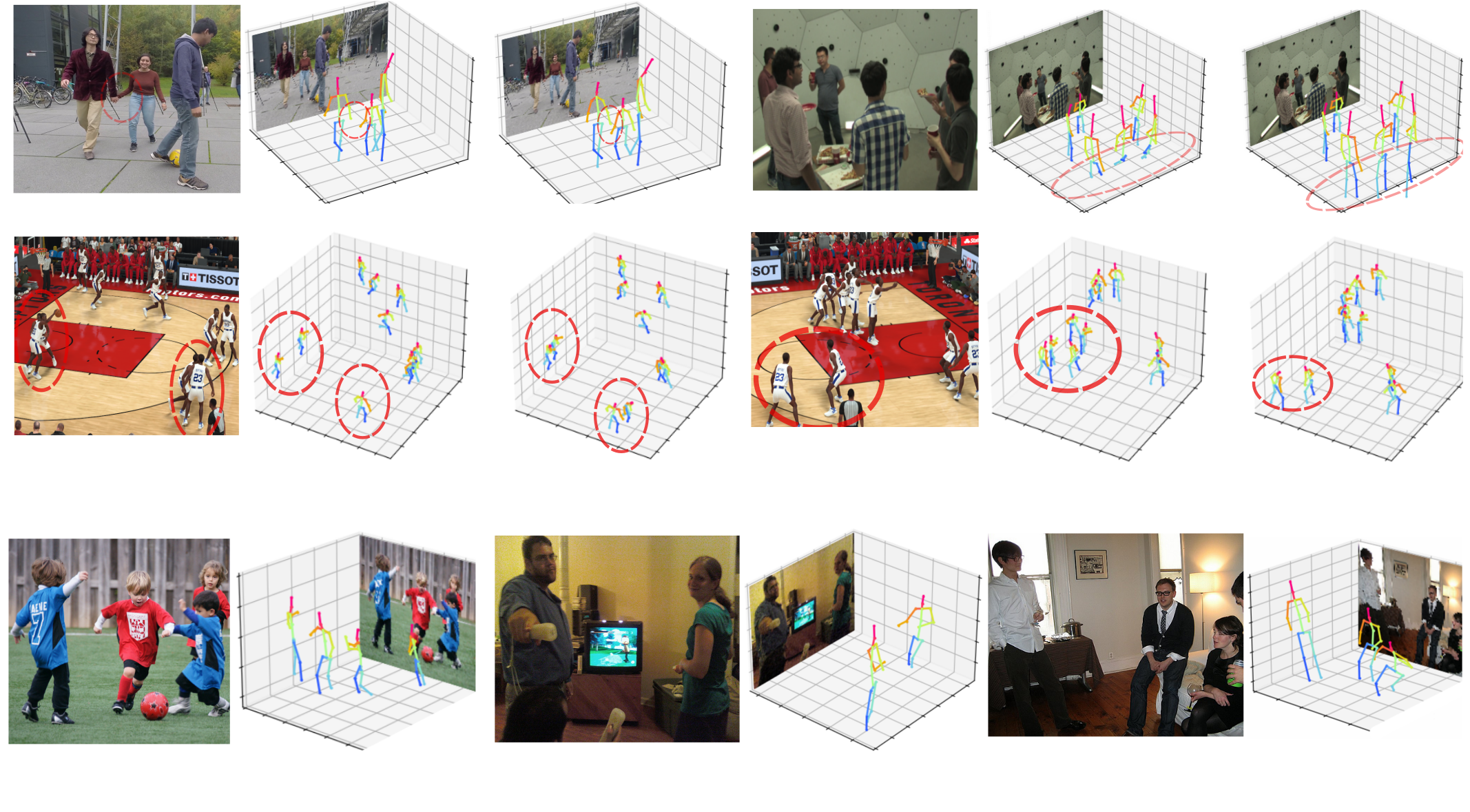}
        \put(-390, 85){{{Input image}}}
        \put(-320, 85){{{SOTA~\cite{Moon_2019_ICCV_3DMPPE}}}}
        \put(-240, 85){{{Ours}}}
        \put(-190, 85){{{Input image}}}
        \put(-120, 85){{{SOTA~\cite{Moon_2019_ICCV_3DMPPE}}}}
        \put(-40, 85){{{Ours}}}
        \put(-390, 0){{{Input image}}}
        \put(-310, 0){{{Ours}}}
        \put(-260, 0){{{Input image}}}
        \put(-170, 0){{{Ours}}}
        \put(-120, 0){{{Input image}}}
        \put(-40, 0){{{Ours}}}
  \vspace{-0.1cm}
  \caption{\small{\textbf{Qualitative results.} We show results on the \textbf{MuPoTS-3D~\cite{singleshot}}, \textbf{Panoptic~\cite{panoptic}}, \textbf{NBA2K~\cite{zhu_2020_eccv_nba}} and \textbf{COCO}~\cite{coco} datasets. Here he show cases of close interactions (first two rows), severe truncation (first row, second column), and our model on images in-the-wild (last row).
  }}
  \label{fig:qualitative}
  \vspace{-0.1cm}
\end{figure*}

We present a direct comparison with the method closest to ours, PI-Net~\cite{guo2021pi}, and show other SOTA methods that also deal with multi-person 3D pose estimation~\cite{HMOR2020,hdnet_2020,smap_eccv2020,zanfir_multipeople_nips18}.
The quantitative results for \textbf{MuPoTS-3D} dataset are reported in Table~\ref{tab:mupots}. Here, we show results of the initialization method RootNet~\cite{Moon_2019_ICCV_3DMPPE} used by both PI-Net~\cite{guo2021pi} and our model. We present, two rows referencing PI-Net~\cite{guo2021pi}. The first one, shows the results when training the model with MuCo-3DHP~\cite{singleshot} dataset, as reported in their work. For a fair comparison, we fine-tune the model with the MuPoTS-3D~\cite{singleshot} dataset and perform the same cross validation. Note that we train and test with MuPoTS-3D due to the reasons explained in Sec.~\ref{sec:implementation}. As it can be seen, our model shows a 3.0\% improvement over this method when estimating the root-relative pose, 2.2\% for AUC for all people and 2.1\% for only matched people. Also, worthy of notice, we remarkably outperform RootNet~\cite{Moon_2019_ICCV_3DMPPE} in all metrics. Comparing to the rest of the methods, our model outperforms the best PCK metric from HDNet by 3.3\%, the best AUC from HMOR by 2.6\%, and the best PCK\textsubscript{abs} from HMOR~\cite{HMOR2020} by 0.3\% . For qualitative comparisons on this dataset, please refer to Fig.~\ref{fig:qualitative}.


\begin{table}[]
\centering
\caption{\ti{Quantitative comparison on the \textbf{MuPoTS-3D}~\cite{singleshot} dataset}. *Results shown for these methods are merely referential as they are not re-trained with the same data as ours. 
Note that both PI-Net and our method use~\cite{Moon_2019_ICCV_3DMPPE} for initialization.}
  \resizebox{0.9 \textwidth}{!}{%
\label{tab:mupots}
\begin{tabular}{@{}lcccccclll@{}}
\toprule
\multicolumn{1}{c}{\multirow{2}{*}{Metric}} &
  \multicolumn{3}{c}{All people} &
  \multicolumn{3}{c}{Matched people} &
  \multicolumn{3}{c}{\multirow{2}{*}{Datasets training}} \\ \cmidrule(lr){2-7}
\multicolumn{1}{c}{} & 3DPCK & 3DPCK abs & AUC rel & 3DPCK & 3DPCK abs & AUC rel & \multicolumn{3}{c}{}                        \\ \midrule
Initial (RootNet~\cite{Moon_2019_ICCV_3DMPPE})    & 81.2  & 31.4      & 39.5    & 82.5  & 32.0      & 40.2    & \multicolumn{3}{l}{MuCo + COCO}             \\
PI-Net~\cite{guo2021pi} (MuCo)        & 82.5  & -         & -       & 83.9  & -         & -       & \multicolumn{3}{l}{MuCo}                    \\
PI-Net~\cite{guo2021pi} (fine-tunned) & 82.8  & -         & 43.9    & 84.3  & -         & 44.7    & \multicolumn{3}{l}{MuPots Cross Validation} \\
\textbf{Ours} &
  \textbf{85.8} &
  \textbf{44.1} &
  \textbf{46.1} &
  \textbf{87.3} &
  \textbf{45.0} &
  \textbf{46.9} &
  \multicolumn{3}{l}{MuPots Cross Validation} \\ \midrule
HMOR~\cite{HMOR2020}*                & 82.0    & 43.8      & 43.5    & -     & -         & -       & \multicolumn{3}{l}{MuCo + COCO +}           \\
HDNet~\cite{hdnet_2020}*               & -     & -         & -       & 83.7  & 35.2      & -       & \multicolumn{3}{l}{MuCo}                    \\
SMAP~\cite{smap_eccv2020}*                & 73.5  & 35.4      & -       & 80.5  & 38.7      & 42.7    & \multicolumn{3}{l}{MuCo + COCO}             \\ \bottomrule
\end{tabular}}
\end{table}

The results for the \textbf{CMU Panoptic} dataset are shown in Table~\ref{tab:panoptic}. We evaluate our method under MPJPE after root alignment following previous works~\cite{zanfir_cvpr_2018_multiple,zanfir_multipeople_nips18}. The dataset presents a challenging scenario as the majority of images contain several people at a time in a closed environment, severely affected by occlusion and truncation. Our method successfully reduces the interference of occlusions and truncation and improves by a large amount the initial estimations (fine-tuned RootNet~\cite{Moon_2019_ICCV_3DMPPE}). To show how our model is able to deal with truncation, results of the metric calculated over all joints in the dataset are presented in Table~\ref{tab:panoptic}, including those that are out of the image and are not visible. Most of these non-visible joint constitute cases of either truncation or occlusion. Our method improves over 30 mm. in average over the initial method. Note that we also have a significant improvement (14.1 mm) over the initial method if we account only for visible joints which is the standard practice. With regards to the SOTA in this dataset (HMOR~\cite{HMOR2020}), we outperform the method by an overall of 5.3 mm and have a consistent improvement over all the actions. For qualitative comparisons on the Panoptic~\cite{panoptic} dataset, please refer to Fig.~\ref{fig:qualitative} and the supplementary material.

\begin{table}
\centering
\caption{\ti{Evaluation on the \textbf{Panoptic}~\cite{panoptic} dataset.} RootNet~\cite{Moon_2019_ICCV_3DMPPE} model was fine-tuned with CMU Panoptic data to provide a better initialization. The reported metric is \textbf{MPJPE} relative to the root joint and results are reported in mm. *The average of~\cite{zanfir_multipeople_nips18} and~\cite{HMOR2020} are recalculated following the standard practice in~\cite{zanfir_cvpr_2018_multiple} and~\cite{smap_eccv2020} (i.e. average over activities) for a more direct comparison.}
\label{tab:panoptic}
  \resizebox{0.6\textwidth}{!}{%
\begin{tabular}{@{}lccccc@{}}
\toprule
\multicolumn{1}{c}{Method}     & Haggling      & Mafia         & Ultim.        & Pizza         & Mean $\downarrow$ \\ \midrule
RootNet~\cite{Moon_2019_ICCV_3DMPPE} (w/ all joints)  & 83.3 & 107.9 & 106.0 & 118.4 & 103.9 \\
Ours (w/ all joints)     & 59.4 & 68.9  & 67.2  & 86.4  & 70.5  \\ \midrule
Zanfir \et~*\cite{zanfir_multipeople_nips18} &72.4	&78.8	&66.8	&94.3	&78.1\\
RootNet~\cite{Moon_2019_ICCV_3DMPPE}  & 52.1 & 65.3  & 58.0  & 80.4  & 63.9  \\
SMAP~\cite{smap_eccv2020}                    & 63.1 & 60.3  & 56.6  & 67.1  & 61.8  \\
HMOR*~\cite{HMOR2020}                     & 50.9 & 50.5  & 50.7  & 68.2  & 55.1  \\
\textbf{Ours} & \textbf{42.0} & \textbf{50.3} & \textbf{47.3} & \textbf{59.4} & \textbf{49.8}     \\\bottomrule
\end{tabular}}
\end{table}


\begin{table}
\centering
\caption{\ti{Evaluation on the \textbf{NBA2K}~\cite{zhu_2020_eccv_nba} dataset}. We use the \textbf{MPJPE} metric. *This method has been fine-tuned with the same dataset and uses the same initial method (RootNet~\cite{Moon_2019_ICCV_3DMPPE}) for fair comparison.}
\label{tab:nba_results}
  \resizebox{0.9 \textwidth}{!}{%
\begin{tabular}{@{}lccccclccccc@{}}
\toprule
\multicolumn{1}{c}{\multirow{2}{*}{Method}} & \multicolumn{5}{c}{MPJPE {[}mm{]}}                         &  & \multicolumn{5}{c}{MPJPE-PA {[}mm{]}}             \\ \cmidrule(lr){2-6} \cmidrule(l){8-12} 
\multicolumn{1}{c}{}                        & Cory  & Glen  & Oscar & Tomas          & Mean $\downarrow$ &  & Cory  & Glen  & Oscar & Tomas & Mean $\downarrow$ \\ \midrule
RootNet~\cite{Moon_2019_ICCV_3DMPPE}                                        & 154.3 & 167.7 & 159.3 & 136.2          & 154.1             &  & 115.8 & 137.5 & 122.4 & 103.9 & 119.7             \\ \midrule
Pi-Net* (fine-tunned)~\cite{guo2021pi}                            & 136.6 & 155.3 & 140.2 & \textbf{119.8} & 137.8             &  & 109.7 & 129.2 & 111.5 & 96.2  & 111.5             \\ \midrule
\textbf{Ours} &
  \textbf{130.0} &
  \textbf{142.0} &
  \textbf{134.7} &
  121.7 &
  \textbf{131.9} &
   &
  \textbf{99.6} &
  \textbf{111.5} &
  \textbf{104.4} &
  \textbf{95.8} &
  \textbf{102.7} \\ \bottomrule
\end{tabular}}
\end{table}

For evaluating on the \textbf{NBA2K dataset}, we use the MPJPE without and with Procrustes Alignment (MPJPE-PA) as shown in Table~\ref{tab:nba_results}. See how both methods that use interaction information from the scene (Pi-Net and ours) are able to improve the results over the initial estimations (RootNet~\cite{Moon_2019_ICCV_3DMPPE}). Furthermore, our method outperforms the baseline (PI-Net~\cite{guo2021pi}) which we fine-tune with this dataset. Moreover, our method shows to be superior than the baseline at capturing interactions and refining the 3D pose of multiple people (see Fig.~\ref{fig:qualitative}). This are particularly interesting results as the dataset contains several people in each scene with high levels of interaction. 






\begin{table}
\centering
\caption{\ti{Importance of different levels of ~\textbf{interaction} in our model.} For each dataset we have 4 different models: the initialization method (RootNet~\cite{Moon_2019_ICCV_3DMPPE}), a no interaction baseline, a baseline with interaction of all ungrouped joints in the scene (scene interaction), and interaction between joints grouped by persons (people interaction). The model that captures interactions between people makes better pose refinements. See text for more details.}
\label{tab:ablation}
  \resizebox{0.6\textwidth}{!}{%
\begin{tabular}{@{}llcc@{}}
\toprule
\multicolumn{1}{c}{Dataset} & \multicolumn{1}{c}{Metric} & MPJPE {[}mm{]} & MPJPE-PA {[}mm{]} \\ \midrule
\multirow{4}{*}{MuPoTS-3D}    & Initial (RootNet)                  & 134.9          & 93.3          \\
                              & Baseline – no interaction          & 136.4          & 95.1          \\
                              & Baseline – scene interaction       & 134.5          & 96.1          \\
                              & \textbf{Ours – people interaction} & \textbf{104.8} & \textbf{79.7} \\ \midrule
\multirow{4}{*}{CMU Panoptic} & Initial (RootNet)                  & 63.9           & 54.6          \\
                              & Baseline – no interaction          & 57.0           & 47.4          \\
                              & Baseline – scene interaction       & 58.2           & 45.1          \\
                              & \textbf{Ours – people interaction} & \textbf{49.8}  & \textbf{40.5} \\ \bottomrule
\end{tabular}}
  \vspace{-0.3cm}
\end{table}
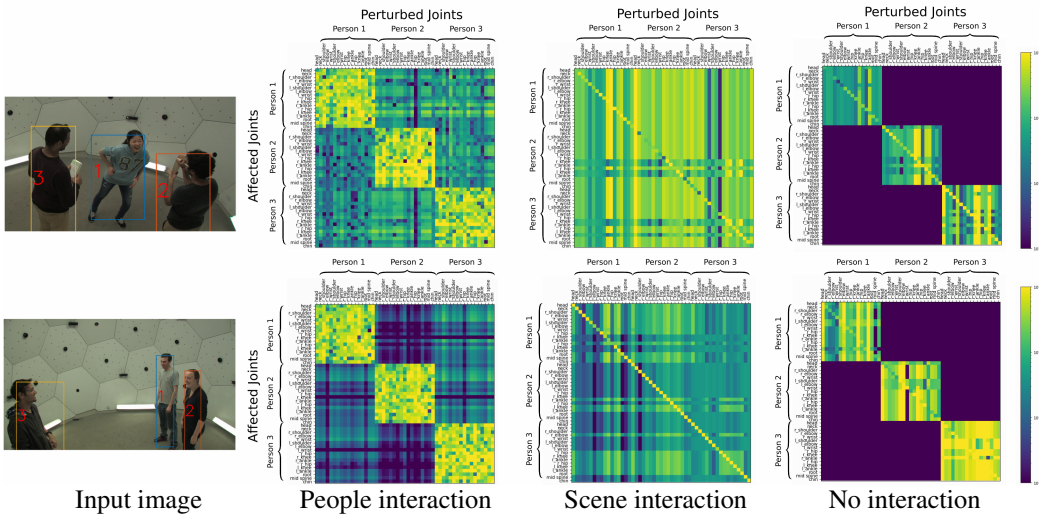
\begin{figure*}[t!]
  \centering
  \includegraphics[width=\linewidth, trim={0.1cm 0.1cm 0cm 0}, clip = true]{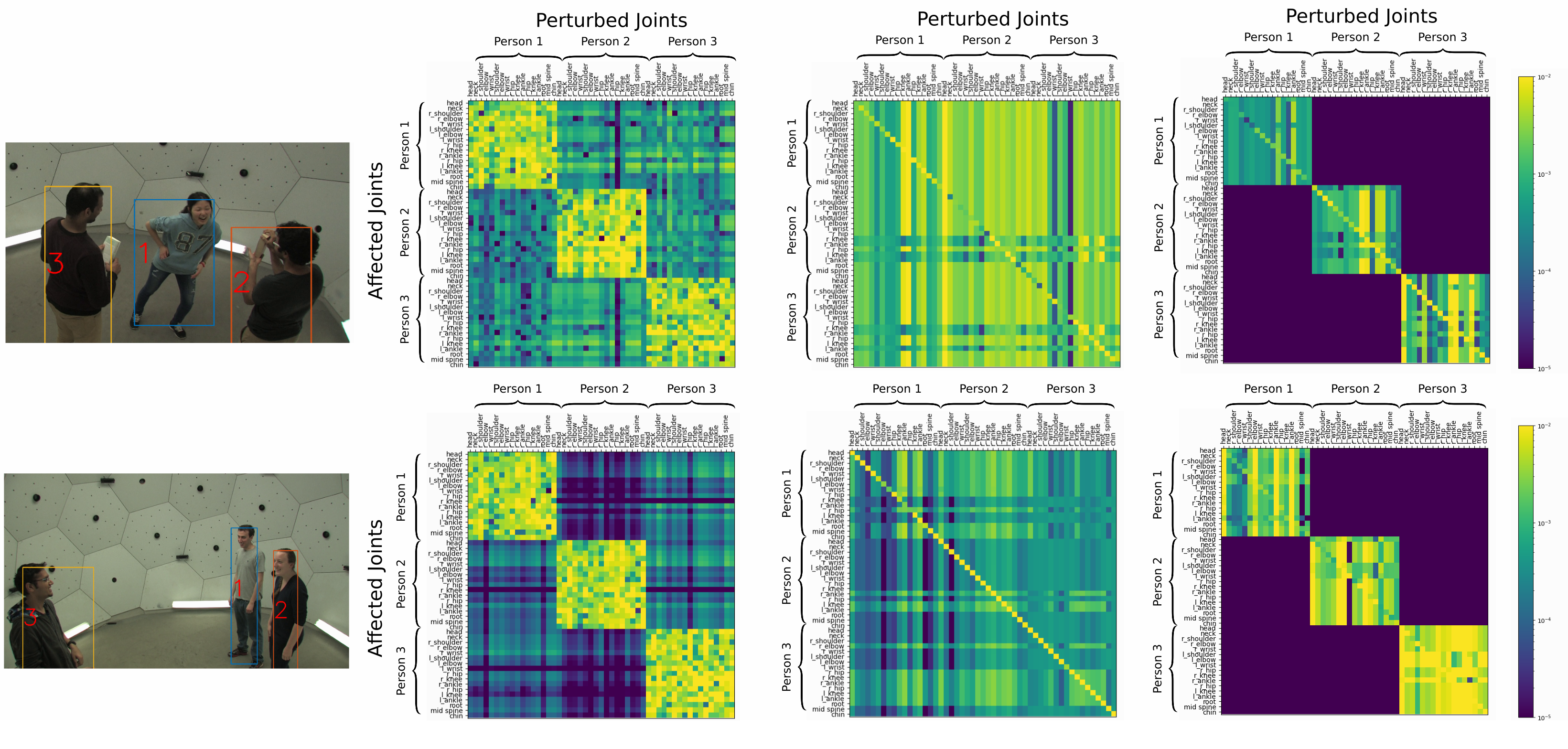}
        \put(-370,-2){{{Input image}}}
        \put(-285,-2){{{People interaction}}}
        \put(-185,-2){{{Scene interaction}}}
        \put(-85,-2){{{No interaction}}}

  \vspace{-0.0cm}
  \caption{\small{\textbf{Interaction analysis.} We show the effect of one joint over all other joints in the scene. Joints are grouped by person. We present 17 joints for each person. Each person's number in the matrices corresponds to the number shown in the bounding box in the images. The magnitude of each matrix element represents the maximum displacement in 3D space measured in meters of the joints in each row caused by the corresponding joint in each column.} 
  }
  \label{fig:interactions}
  \vspace{-0.1cm}
\end{figure*}

\subsection{Interaction vs. no interaction}
Having shown the effectiveness of our method at refining 3D poses, we continue with a careful analysis of our interaction component. Table~\ref{tab:ablation} shows how the level at which we enforce the interaction to be learned affects the performance in comparison to the initial estimations. We define three different levels of interaction: (1) no interaction, (2) scene interaction, and (3) people interaction. The latter corresponds to our final method. For all the cases, we use the same architecture. However, we change the interaction levels by changing what we input to our method. To eliminate learning interactions ({\em no interaction}), we input each person's pose individually as a unique and different set and not together. In this manner, it is impossible for the model to build an interaction-based embedding. At most, the model remains restricted to capture self-joint interactions. To enforce learning what we refer to as \ti{scene interaction}, we make each joint in the scene a set by itself. Having each joint as set element instead of the whole person's pose, we enforce a representation that can learn interactions between joints but without knowing which joint corresponds to which person. Thus, loosing the sense of person as an "entity". The results from Table~\ref{tab:ablation} show that our method based on people's interaction clearly outperforms other degrees of interaction consistently over both datasets. We report the results on the MuPoTS-3D and Panoptic datasets and use the MPJPE metric only for simplicity. 

Figure~\ref{fig:interactions} gives us insight on what happens with the refinement on each level of interaction. Here, we present two images containing three persons each: high interaction between individuals (top image), low interaction (bottom image). Additionally, for each image we show three matrices, each regarding an interaction type presented above. Each matrix depicts the effect of the pose refinement when perturbing one joint over all of the joints of every person in the scene. This also includes the person whose joint has been perturbed. Each column represents the joint being perturbed and each row represents the affected joints. The perturbation applied to the joints is a displacement in the positive directions of x, y and z in the 3D space by 10 cm. The magnitude in each element in the matrix represents the maximum absolute value of the change in any of the 3D space coordinate direction in meters. With this setup, we can study how one person's joint affects other person's joints as well as its own. Here, we notice some key observations. (1) for the cases in which we enforce to learn interactions (people and scene), the effect of one person's joint over other person's joints (effect of interaction) is higher for the image on the top (high interaction) than for the one in the bottom (lower interaction). (2) in the case of people interaction, it can be clearly seen that one person's joint affects in greater degree the pose of its own body in contrast to other people's body. This is expected, as the model here has the notion of which joints correspond to which person. This does not happen in the scene interaction case. Also, (3) we can confirm that in the case of no-interaction, one person's pose has no effect over others. The reader is referred to the supplemental material for additional examples.

\begin{figure*}[t!]
  \includegraphics[width=\linewidth, trim={0.1cm 0.1cm 0cm 0}, clip = true]{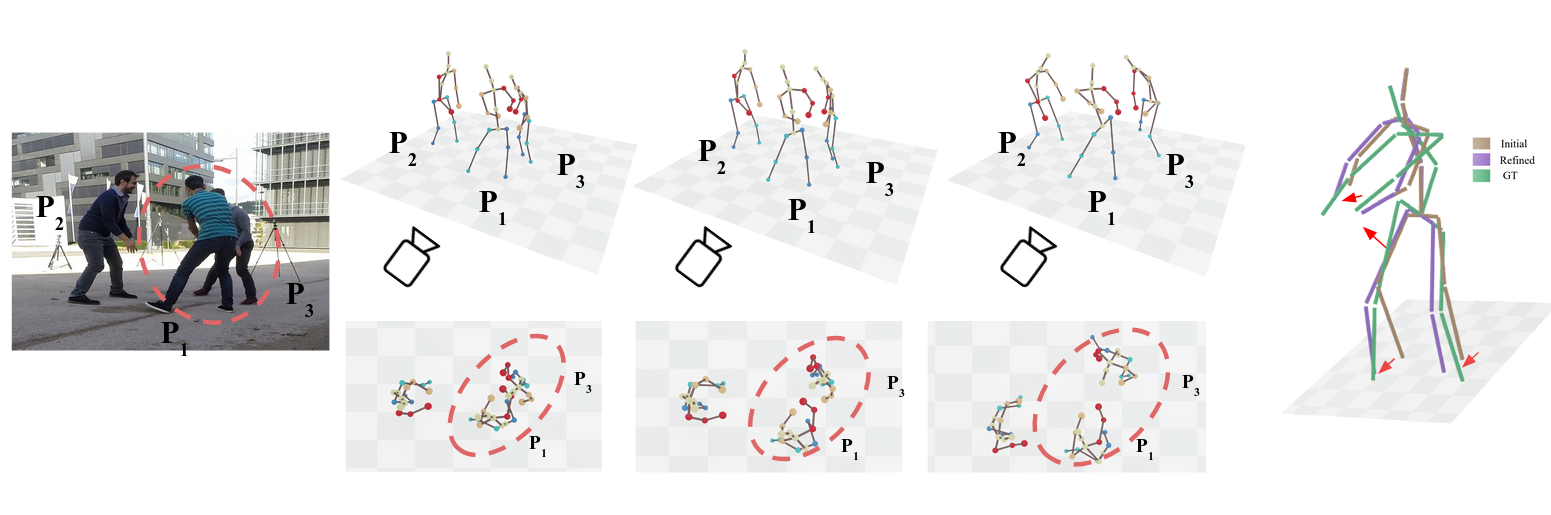}
        \small{\put(-380,-2){{{Input image}}}}
        \small{\small{\put(-292,125){{{Side-view}}}}}
        \small{\small{\put(-290,51){{{Bird-view}}}}}
        \small{\put(-308,-2){{{Initial Estimation}}}}
        \small{\small{\put(-215,125){{{Side-view}}}}}
        \small{\small{\put(-215,51){{{Bird-view}}}}}
        \small{\put(-225,-2){{{Refined Poses}}}}
        \small{\small{\put(-139,125){{{Side-view}}}}}
        \small{\small{\put(-139,51){{{Bird-view}}}}}
        \put(-150,-2){{{Ground Truth}}}
        \put(-80,-2){{{Overlay poses for P\textsubscript{1}}}}
        
  \centering
  \vspace{-0.0cm}
  \caption{\small{\textbf{Effects of the pose refinement.} From left to right: Input image, initial 3D pose estimations, refined poses, ground truth, and detail of the update on person P$_1$'s joints. For each estimation we include a bird-view so that absolute translation is better appreciated. The last column shows the root-relative pose improvement.} 
  }
  \label{fig:explain}
  \vspace{-0.3cm}
\end{figure*}
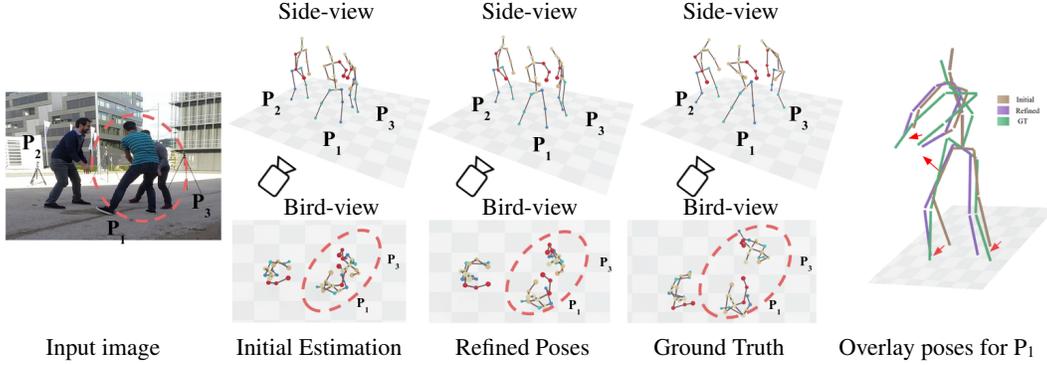

\subsection{Effects of the refinement over initial estimations}
We show the effect of our model in refining the initial estimations. Our method can improve both absolute and root-relative poses while more effectively dealing with inter-person occlusions and truncations. This is achieved because our \textit{interaction embedding} enables the model to reason directly in the 3D space, whereas other methods can only reason from 2D image cues. Furthermore, our loss encourages the model to learn a refinement for both the absolute and the root-relative poses. See Fig.~\ref{fig:explain}. The three middle columns depict the initial estimation, the refined poses and the ground truth from a slightly rotated camera view along with a bird-view, respectively. From these views, we can appreciate the interaction between person 1 (P\textsubscript{1}) and person 3 (P\textsubscript{3}). The initial estimation does not take into account this interaction and, therefore, makes the mistake of overlapping the two bodies. Our model yields to more realistic estimations by exploiting these interactions. The rightmost picture shows the initial, refined and ground truth root-relative poses for P\textsubscript{1}. See how our estimations correct the initial joint positions taking them closer to the ground truth (highlighted in red arrows). For example, the hands and ankle joints are closer to the ground truth than the initial estimations. The same happens with the hip joints. Additionally, Fig.~\ref{fig:teaser} shows the input image, the ground truth and the refined poses overlapped with the initial estimations (in transparency) both in a tilted camera view and a bird-view. Here, we can also appreciate an interaction between the people that are about to hug (P\textsubscript{2} and P\textsubscript{3}). Better seen in the bird-view, our refinement locates both persons in a more coherent way, whereas, the initial estimation places them further apart.

\subsection{Qualitative Results}

Qualitative results on the MuPoTS-3D, Panoptic, NBA2K ana COCO datasets are shown in Fig.~\ref{fig:qualitative}. Row 1 fist column shows a case where people are closely interacting with each other (holding hands). Our model corrects the persons poses so their hands are closer together. Row 1 second column shows how our method corrects cases of severe truncation. Row 2 shows results on the NBA2K~\cite{zhu_2020_eccv_nba} dataset and the last row shows as well results on images-in-the-wild from COCO~\cite{coco} dataset. NBA2K~\cite{zhu_2020_eccv_nba} 
presents several interactions in each scene. We show how our method improves over the SOTA~\cite{Moon_2019_ICCV_3DMPPE}, especially in cases were people need to be grouped closely together. Our method captures the interactions between the players and can determine which players should be grouped together. In each case, with dotted red circles, we show either a correctly located group of players (refined poses) or incorrectly placed players (initial poses).

\subsection{Computational complexity}
\label{sec:complexity}
Our model is fast and has very few parameters. Also, we need less data to train to make the refinements in direct comparison to~\cite{guo2021pi} and indirectly compared to~\cite{Moon_2019_ICCV_3DMPPE,HMOR2020,smap_eccv2020}. We use 90K model parameters, need 5.1M FLOPS to make an inference which takes 2ms and we required roughly 15K training samples. In comparison to~\cite{Moon_2019_ICCV_3DMPPE}, which takes 250.7ms to make an inference, using our model as a refinement step constitutes a negligible overhead. For a more comprehensible comparison table, please refer to our supplementary material.



\section{Conclusions}\label{sec:conclusion}

In this paper we have proposed a novel algorithm to tackle the problem of multi-person 3D pose estimation from one single image. Building on the Set Transformer paradigm, we have introduced a holistic encoding of the entire scene, given an initial set of potentially noisy input 3D body poses. This encoding captures multi-person relationships, does not depend on the input order and can represent an arbitrarily large number of inputs. We use it to refine the initial poses in a residual manner. A thorough evaluation shows that our approach provides state-of-the-art results on several benchmarks.
Additionally, the proposed module is computationally efficient and can be used as a post-processing step for any 3D pose detector in multi-people scenes to improve its accuracy and make it more robust to truncation and occlusions. 



\bibliography{references}

\end{document}